\documentclass[letterpaper, 10 pt, journal, twoside]{IEEEtran}
\usepackage{amsmath,amsfonts}
\usepackage{algorithmic}
\usepackage{array}
\usepackage[caption=false,font=normalsize,labelfont=sf,textfont=sf]{subfig}
\usepackage{textcomp}
\usepackage{stfloats}
\usepackage{url}
\usepackage{verbatim}
\usepackage{graphicx}
\usepackage{makecell}
\usepackage{xcolor}
\def\BibTeX{{\rm B\kern-.05em{\sc i\kern-.025em b}\kern-.08em
    T\kern-.1667em\lower.7ex\hbox{E}\kern-.125emX}}
\usepackage{balance}

\title{
A Highly Maneuverable Flying Squirrel Drone \\ With Agility-Improving Foldable Wings
}

\author{Dohyeon~Lee$^{1*}$, Jun-Gill~Kang$^{2*}$, and~Soohee~Han$^{1**}$,~\IEEEmembership{Senior~Member,~IEEE}%

\thanks{
$^*$Dohyeon Lee and Jun-Gill Kang are co-first authors. $^{**}$Corresponding author: Soohee Han.
} 
\thanks{$^{1}$Dohyeon Lee and Soohee Han are with Department of Convergence IT Engineering, Pohang University of Science and Technology (POSTECH), 37673 Pohang, South Korea.
        {\tt\footnotesize \{dohyeon, soohee.han\}@postech.ac.kr}}%
\thanks{$^{2}$Jun-Gill Kang is with Defence AI Center, Agency for Defense Development(ADD), Daejeon, 34186, South Korea.
        {\tt\footnotesize jungillkang@gmail.com}}%
}

\begin{document}
\maketitle
\begin{abstract}
Drones, like most airborne aerial vehicles, face inherent disadvantages in achieving agile flight due to their limited thrust capabilities. These physical constraints cannot be fully addressed through advancements in control algorithms alone. Drawing inspiration from the winged flying squirrel, this paper proposes a highly maneuverable drone with agility-enhancing foldable wings. The additional air resistance generated by appropriately deploying these wings significantly improves the tracking performance of the proposed ``flying squirrel" drone. By leveraging collaborative control between the conventional propeller system and the foldable wings—coordinated through the Thrust-Wing Coordination Control (TWCC) framework—the controllable acceleration set is expanded, allowing for the production of abrupt vertical forces unachievable with traditional wingless drones. The complex aerodynamics of the foldable wings are captured using a physics-assisted recurrent neural network (paRNN), which calibrates the angle of attack (AOA) to align with the real-world aerodynamic behavior of the wings. The model is trained on real-world flight data and incorporates flat-plate aerodynamic principles. Experimental results demonstrate that the proposed flying squirrel drone achieves a 13.1$\%$ improvement in tracking performance, as measured by root mean square error (RMSE), compared to a conventional wingless drone.
A demonstration video is available on YouTube: https://youtu.be/O8nrip18azY.
\end{abstract}
\begin{IEEEkeywords}
Biomimetics, Aerial Systems: Mechanics and Control, AI-Based Methods.
\end{IEEEkeywords}
\section{Introduction}
\IEEEPARstart{D}{rones} are widely utilized in diverse applications such as inspection \cite{pfreundschuh2023resilient}, data acquisition \cite{yoon2024spinning}, and rescue operations \cite{schedl2021autonomous}, where time is often critical and necessitates high-speed performance, especially in hard-to-reach or hazardous locations.

However, drones face inherent limitations, such as underactuated dynamics and complex aerodynamics, which hinder agile flight in confined spaces and rapid emergency response without collisions or instability. For this reason, there have been various control strategies  such as trajectory optimization \cite{oleynikova2018safe, liu2018convex}, neural network-based control \cite{li2017deep}, conventional PID control, geometric control \cite{lee2010geometric}, differential flatness-based control \cite{sreenath2013geometric}, and integral backstepping control. However, the drone's attitude controller alone cannot overcome its own physical constraints.

For additional control effort, the same authors suggested a flying squirrel-inspired drone with controllable foldable wings that enhance flight performance by generating air resistance \cite{kang2023highly}. This previous work addressed predefined paths in controlled indoor settings, limiting its applicability in real-world scenarios.

To address the aforementioned existing limitations, the authors propose a more general trajectory tracking algorithm with an automatic wing controller. For attitude control, an integral backstepping approach is adopted due to its low computational cost and inherent guaranteed stability, which aligns the goal of developing an onboard-controlled, flying squirrel-inspired drone. In addition, we employ a neural network-based wing controller combined with a PID-based position controller to enhance trajectory tracking capabilities. By integrating both strategies, the proposed approach is suitable for agile flight even in non-differentiable trajectories such as sharp turns.


Our main contributions are twofold. First, we introduce a noise robust, physics-assisted recurrent neural network (paRNN) to learn the aerodynamics of the foldable wings in sample-efficient way by supervised learning. Second, we propose a sophisticated model-based control system, the Thrust-Wing Coordination Control (TWCC) algorithm, which leverages the learned wing dynamics to optimize the timing of folding and unfolding without compromising stability criteria. By combining precise aerodynamic modeling with an advanced control algorithm, the proposed system achieves high tracking performance across diverse and unpredictable trajectories, enabling the drone to operate effectively in outdoor environments.

To validate the proposed dynamics learning algorithm, we compare the learned paRNN against a vanilla RNN using four previously unseen flying datasets to predict the aerodynamics forces. paRNN achieves at least a 24.4$\%$ improvement (up to 70.4 $\%$) in estimation performance, as measured by root mean square error (RMSE).

To validate the proposed tracking controller, we conduct outdoor trajectory tracking experiments while avoiding virtual obstacles. Quantitatively, results show that the proposed flying squirrel drone achieves a 13.1 $\%$ improvement in tracking performance, measured by  RMSE, compared to a conventional wingless one.

This letter is organized as follows: Section II outlines the hardware configuration, mathematical modeling, and sensors of the drone. Section III provides a detailed explanation of TWCC. Section IV discusses the estimation of air resistance from silicone wings using paRNN. Section V validates our method through simulation and experimentation. Finally, Section VI provides a summary of the contributions of this work along with future research directions.

\begin{figure*}[t]
    \centering
    \includegraphics[scale=0.45]{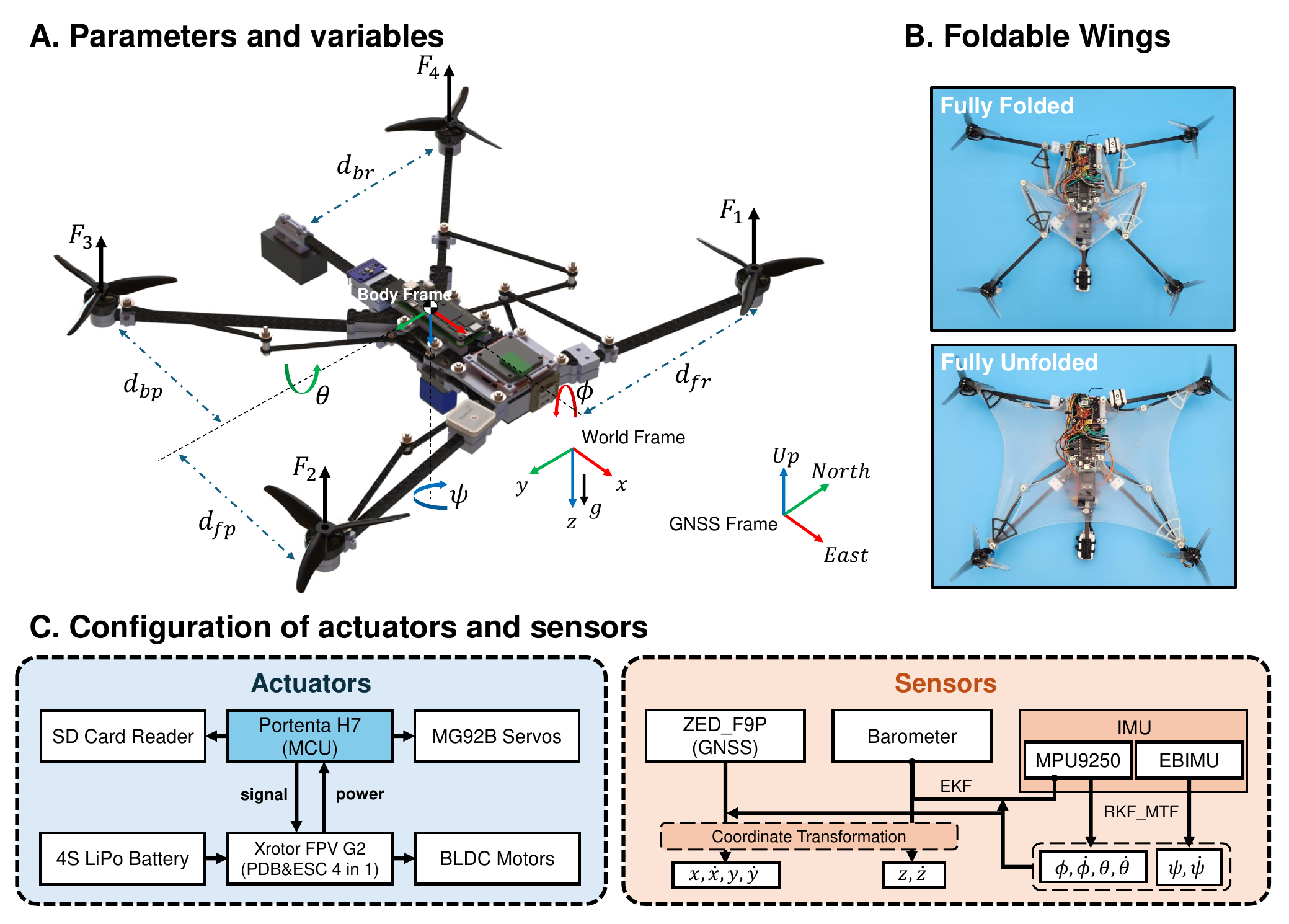}
    \caption{Overview of the flying squirrel drone. The frames used (Body, World, GNSS) can be found in A, and the actual image of the drone can be seen in B.}
    \label{silicone_wing}
\end{figure*}

\section{ Overview of the Flying Squirrel Drone \label{section2} }
\subsection{Specifications of the Flying Squirrel Drone} 
Table I shows the overall specifications of the flying squirrel drone used in this work. A custom, lab-built flight controller implemented on a micro controller unit (MCU) coordinates the propellers and wings. The main body and the crank arms that fold or unfold the wings are made of carbon. The wings, made of silicone, weigh only 24 g, less than 5 percent of the total. The crank mechanism and the fabrication process of silicone wings are detailed in \cite{kang2023highly}.

\begin{table}[ht]
\caption{Specifications of a flying squirrel drone}
\label{table_spec}
\begin{center}

\begin{tabular}{|c|c|}
\hline
\textbf{Parameters /} & \textbf{Values /} \\
\textbf{Components } & \textbf{Quantities} 
\\
\hline
\makecell{Size } & \makecell{276.06~mm $\times$ 373.2~mm $\times$ 25~mm \\ (L $\times$ W $\times$ H) \\} \\
\hline
Mass & 0.548~kg (0.024~kg for silicone wings) \\
\hline
Moments of Inertia & \makecell{$I_{xx} =  3.22\times10^{-3}{\rm{kgm^2}}$ \\ $I_{yy} =  4.68\times10^{-3}{\rm{kgm^2}}$ \\ $I_{zz} = 7.72\times10^{-3}{\rm{kgm^2}}$} \\
\hline
MCU & 1 $\times$ STM32H747 Cortex M7/M4 (dualcore) \\
\hline
Power source & 1 $\times$ 14.8~V, 75C LiPo battery \\
\hline
Motors & \makecell{4 $\times$ T-motor 2004(3000 KV) \\ 2 $\times$ MG92B servo} \\
\hline
Sensors & \makecell{2 $\times$ IMU, 1 $\times$ Barometer, 1 $\times$ GNSS} \\
\hline
\end{tabular}
\end{center}
\end{table}
\subsection{System Modeling}
The force control input can be generated from the desired movement-inducing torques and thrust as follows:
\begin{equation}
\begin{aligned}\label{Integral Backstepping_translate_1}    
    &\textbf{F} = \textbf{T}^{-1}\textbf{U} 
\end{aligned}
\end{equation}
where $\textbf{U}$, $\textbf{T}$, and $\textbf{F}$ are given by
\begin{equation}
\begin{aligned}\label{Integral Backstepping_translate_2}
\scriptstyle
    \textbf{U} = \left [ \begin{matrix} 
    U_{\Sigma} \\
    U_{\phi} \\ 
    U_\theta \\ 
    U_\psi \end{matrix}
    \right ], 
    \textbf{T} = \left [ \begin{matrix} 
    1&1&1&1 \\
    d_{fr}&d_{br}&-d_{br}&-d_{fr} \\ 
    d_{fp}&-d_{bp}&-d_{bp}&d_{fr} \\ 
    -ct_1&ct_2&-ct_1&ct_2\end{matrix}
    \right ], 
    \textbf{F} = \left [ \begin{matrix} 
    F_1 \\
    F_2 \\ 
    F_3 \\ 
    F_4 \end{matrix}
    \right ]
\end{aligned}
\end{equation}
$U_{\Sigma}$ is the sum of the thrust of the four rotors which controls the altitude, $U_{\phi,\theta,\psi}$ represent torques, $ct_1$ and $ct_2$ are the ratios of torque to thrust produced by the propeller when the rotor rotates in the clockwise and counterclockwise directions, respectively, and all other parameters and variables are specified in Fig. 1 and Table I.
Note that actual physical control inputs are generated in the form of PWM signals. Accordingly, through experiments, the voltage corresponding to the PWM signal is fitted to the motor thrust by a 3rd order polynomial.
The dynamics of a flying squirrel drone can be written as follows:
\begin{align}
\begin{split}\label{equation of motion}
    \ddot{\mathbf{r}} &= g \mathbf{e}_3 + \frac{1}{m} R_{wb} \mathbf{f} + \frac{\bold{f_a}}{m}
\\
    \dot{\boldsymbol{\omega}} &= J^{-1} (-\boldsymbol{\omega} \times J \boldsymbol{\omega} + \boldsymbol{\tau}) 
\end{split}
\end{align}
where $\mathbf{r} \in \mathbb{R}^3$ is the position vector in the world frame in Fig. \ref{APSC_structure}, $g$ is the gravitational acceleration, $\mathbf{e}_3$ is a $z$ directional unit vector (heading down), $m$ is the drone mass, $\mathbf{f}=\begin{bmatrix} 0 & 0 & -U_{\Sigma} \end{bmatrix}^{T}$ is the total thrust, $\mathbf{f_a} = \begin{bmatrix} f_{a,x} & f_{a,y} & f_{a,z} \end{bmatrix}^{T}$ is the world frame aerodynamic force, $R_{wb} \in SO(3)$ is the rotation matrix from the world frame to the body frame, $\boldsymbol{\omega} \in \mathbb{R}^3$ is the angular velocity in the body frame, $J$ is the moment of inertia matrix, and $\boldsymbol{\tau}=\begin{bmatrix} U_{\phi} & U_{\theta} & U_{\psi} \end{bmatrix}^{T}$ represents the control torque input.
According to the flat wing theory \cite{roberts2009controllability}, the flat plate experiences the following aerodynamic force:
\begin{gather}\label{flat theory in 3D_1}
\mathbf{f}_{flat}(\phi,\theta,\psi,v_x,v_y,v_z) = -\rho_{air} \sin\alpha A\left\| \mathbf{v} \right\|^2 \cdot \mathbf{n}
\end{gather}
where $\alpha$ is the angle of attack, defined as the angle between the air flow’s direction and the wing surface; $A$ is the area of the plate; $\mathbf{v}$ is velocity of drone; and $\mathbf{n}$ is given by 
\begin{gather}\label{flat theory in 3D_3}
\mathbf{n} = R_{wb}\mathbf{e_3} 
= \begin{bmatrix} \sin \psi \sin\phi + \cos\psi \sin\theta \cos\phi \\
-\cos\psi \sin\phi + \sin\psi \sin\theta \cos\phi \\
\cos\theta \cos\phi \end{bmatrix} 
\end{gather}
It should be noted that the aerodynamic force in (\ref{flat theory in 3D_1}) includes both lift and drag forces. The angle of attack, $\alpha$, will be adjusted through the paRNN, as discussed later in this paper.

\subsection{Online Attitude and Position Estimation}

The state of the drone consists of its attitude and position, both of which must be estimated accurately for effective control. Attitude estimation is typically achieved through sensor fusion, combining data from accelerometer, magnetometer, and gyroscope \cite{jing2017attitude,de2011uav,abdelfatah2021uav,liu2014auto}.
For position control, the drone in this study utilizes GNSS, IMU, and barometer. The $x$-axis and $y$-axis directional estimations are achieved through filtering within the GNSS module, while the $z$-axis position is estimated using a Kalman filter that combines the $z$-axis acceleration from the IMU with the readings from the barometer.
To evaluate the effect of the wing membrane on the drone, we conducted experiments under high-speed and rapid-acceleration conditions, where state estimation accuracy significantly deteriorated. To address this issue, we applied the RKF-MTF algorithm \cite{candan2021robust}, which adaptively adjusts the associated covariance based on external acceleration, thereby preserving high estimation accuracy.

\section{Controller Design \label{section3}}

In this paper, the control objectives considered for a flying squirrel drone are as follows:
\\
\begin{itemize}
\item High tracking performance in agile flight.
\item  Real-time optimal decision-making on when to fold or unfold wings. 
\item Timely application of feedforward aerodynamic control forces.
\end{itemize}
~\\
To achieve these objectives, we propose Thrust-Wing Coordination Control (TWCC), a control system that assesses the effectiveness of aerodynamic forces and overrides the conventional attitude and position controllers without compromising stability criteria.



\subsection{Integral Backstepping Control}

As in our previous work \cite{kang2023highly}, an integral backstepping controller, a nonlinear controller that satisfies Lyapunov stability criteria, was used for attitude control. On our hardware, the attitude controller operates at 300 Hz.
Information on the application of integral backstepping control can be found in \cite{bouabdallah2007full}.

\subsection{Thrust-Wing Coordination Control (TWCC)}
\begin{figure*}[t]
    \centering
    \includegraphics[width=\linewidth]{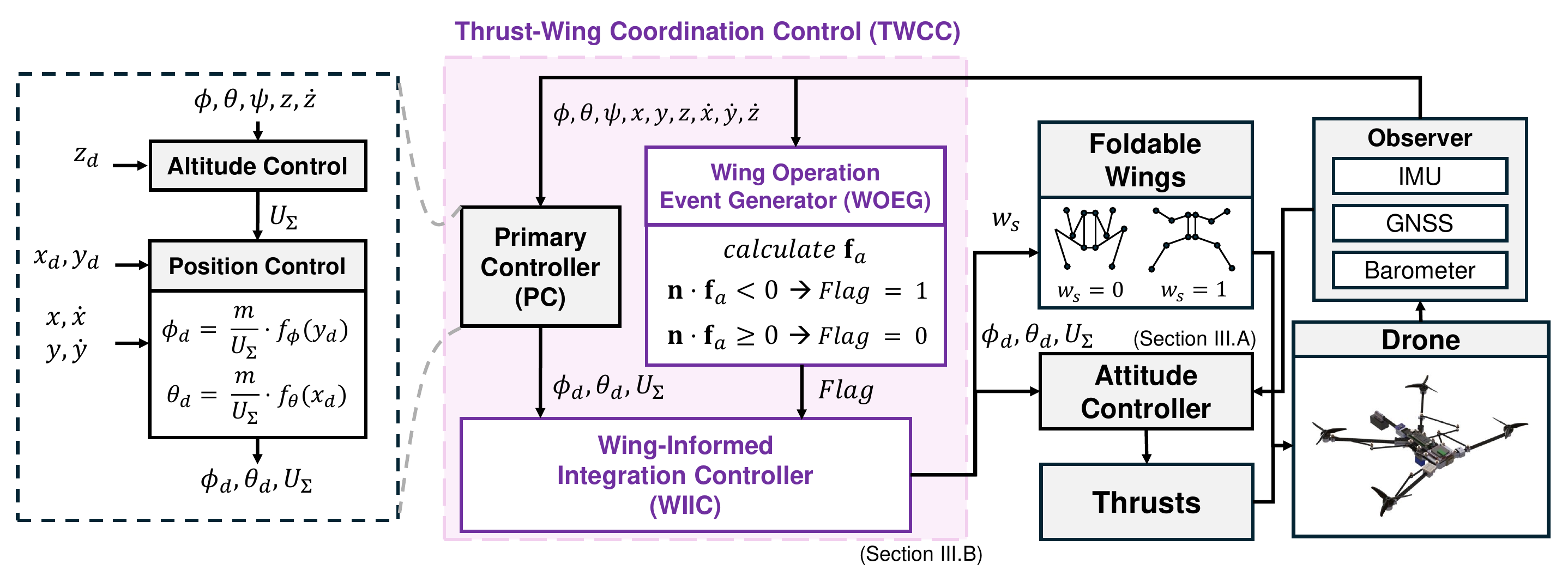}
    \caption{Overview of TWCC. The structural diagram illustrates how the components of TWCC—PC, WOEG, and WIIC—operate. As shown in the figure, details related to TWCC can be found in Section III.B, while information on the Attitude Controller is provided in Section III.A.}
    \label{APSC_structure}
\end{figure*}
As diagrammed in the leftmost section of Fig. \ref{APSC_structure}, a general drone controller first determines $U_{\Sigma}$ through altitude control and subsequently performs the corresponding position control \cite{bouabdallah2007full}.
This type of control performs well in general, non-agile flight without highly maneuverable operations. However, this approach leads to serious tracking performance degradation for highly agile flight scenarios, such as sudden changes in direction. For example, sudden appearance of an obstacle generates reference trajectories requiring large roll and pitch angles. This can easily cause control saturation. Moreover, physical limitations on achievable roll and pitch angles may prevent the system from achieving these large angles in practice.
To overcome this problem, we propose TWCC, which optimally utilizes thrust and foldable wings while considering physically allowable roll and pitch angles.
As seen in Fig. \ref{APSC_structure}, TWCC consists of a wing operation event generator (WOEG), a wing-informed integration controller (WIIC), and a primary controller (PC). 
A brief summary of the operation of each module in TWCC is as follows:
\begin{itemize}
\item If $\phi_d$ and $\theta_d$ have not reached their maximum angles, and the desired thrust $U_{\Sigma}$ does not exceed its limit, then $\phi_d$, $\theta_d$, and $U_{\Sigma}$ are determined by the PC.
\item WOEG evaluates whether the estimated aerodynamic force is beneficial and sets $Flag = 1$ if spreading the wings is favorable.
\item If $\phi_d$ or $\theta_d$ reaches its maximum angle and $Flag = 1$, the wings are spread, and $\phi_d$, $\theta_d$, and $U_{\Sigma}$ are determined by the WIIC.
\end{itemize}
The parameter $Flag$ indicates whether the wings should be spread, as determined by WOEG. WOEG compares the drone body's normal vector with estimated aerodynamic force vector to determine whether spreading the wings will assist or hinder the drone’s movement. When additional force is required, the controller switches from PC to WIIC, which activates the wings. While WIIC controls the system, PC continues to operate internally, evaluating transition conditions. Details of the PC, WOEG, and WIIC are provided in subsequent sections.




\vspace{0.6em}\noindent\textbf{Primary Controller (PC).} For position control in PC, a widely employed cascade PID control method is adopted. Each axis is controlled as follows: 
\begin{gather}
\begin{split}
&{a}_{cmd,j} = k_{p,j}^v e_j + k_{i,j}^v \int e_j dt + k_{d,j}^v \dot{e}_j \nonumber 
\end{split}
\\
\begin{split}
&e_j = v_{jd}  - v_j  
\end{split}
\\
\begin{split}
&v_{jd} = k_{p,j}^p (j_d - j)  \nonumber 
\end{split}
\end{gather}
where \( j \in \{x, y, z\} \), and the subscript \( d \) denotes ``desired". The parameters \( k_p^v, k_i^v, k_d^v \) represent the PID gains for velocity control, while \( k_p^p \) is the proportional gain for position control. The desired velocity is obtained through a position-based P controller, and it is passed to the velocity PID controller to ultimately determine the final acceleration command. The terms \( \phi_d, ~\theta_d,\) and\(~ U_{\Sigma} \), as shown in Fig. \ref{APSC_structure}, are given by:
\begin{align}
U_{\Sigma} &= \frac{m}{ \cos\phi \cos\theta}(g-{a}_{cmd,z}) \nonumber\\
\theta_d &= - \frac{m}{U_{\Sigma}} a_{cmd,x} \\
\phi_d &= \frac{m}{U_{\Sigma}} a_{cmd,y} \nonumber
\end{align}
\noindent\textbf{Wing Operation Event Generator (WOEG).} WOEG determines whether the generated aerodynamic force acts favorably or adversely on the desired translational motion of a drone when the wings are spread. WOEG first estimates the aerodynamic force and then compared with the normal vector of a drone body to assess its usefulness. This entire process is illustrated in Fig. \ref{APSC_structure}. The aerodynamic force $\mathbf{f}_{a}$ can be calculated in two methods: the first applies the aforementioned flat-wing theory, while the second employs a neural network-based approach, which is discussed in the next section.
If WOEG outputs $Flag = 1$, and PC requests a roll or pitch greater than their maxima $\phi_{\max}$ and $\theta_{\max}$, WIIC takes over the control decision from PC.
In general, the actual acceleration of a conventional wingless drone, \( a_{act,x} \) and \( a_{act,y} \), can be approximated as functions of the commanded acceleration \( a_{cmd,x} \) and \( a_{cmd,y} \) from a general controller, such as the PC.
The attitude angles of an actual drone are limited by physical constraints and stability considerations, i.e., \( \underline{\phi} < \phi < \overline{\phi},~~\underline{\theta} < \theta < \overline{\theta} \). 
If the controller requests attitude angles beyond these limits, the desired steering is not achieved. Likewise, since \( U_{\Sigma} \) is bounded, the drone fails to reach \( a_{cmd} \) if the required thrust exceeds \( \overline{U_{\Sigma}} \).  As a result, the actual acceleration \( a_{act} \) is lower than the commanded one \( a_{cmd} \) from PC when the control input is saturated.
For the \( y \)-axis, the actual acceleration \( a_{act,y} \) falls short of the commanded \( a_{cmd,y} = \phi_d \frac{U_{\Sigma}}{m} \) when \( \phi_d > \overline{\phi} \) or \( U_{\Sigma} > \overline{U_{\Sigma}} \). The same applies to the \( x \)-axis when corresponding control input falls below its minimum limit.
\noindent \textbf{Wing-Informed Integration Controller (WIIC).} WIIC is triggered when the drone's controller requests attitude angles beyond their allowable ranges while $Flag=1$. 
Specifically, the roll or pitch component that reaches its limit is fixed at the constrained value. Instead of adjusting the angles, \( U_{\Sigma} \) is regulated to achieve the intended motion with spreading wings (\( w_s = 1 \)).
For example, if the pitch angle $\theta$ is out its allowable range, the following thrusts and angles are generated:
\begin{align}\label{AEC:unfoldstate}
\begin{split}
\theta_d &= \theta_{\max} \\ 
U_{\Sigma} &= -\frac{m}{\theta_{\max}}(a_{cmd,x} - \frac{f_{a,x}}{m}) \\ 
\phi_d &= \frac{m}{U_{\Sigma}} (a_{cmd,y}-\frac{f_{a,y}}{m})
\end{split}
\end{align}
$f_{a,x}$ and $f_{a,y}$ represent the air resistance generated by the spread wings, which not only reduces strain on the drone motor but also increases the available range of the $U_{\Sigma}$, enabling it to generate a stronger force in the desired direction. When the motor thrust is limited, insufficient thrust can be adequately compensated by the aerodynamic forces $f_{a,x}$ and $f_{a,y}$ in (\ref{equation of motion}). The same process applies if the roll angle of a drone reaches its limit as in the pitch one above. 
If none of the conditions to be controlled by WIIC are satisfied, WIIC folds the wings ($w_s = 0$), and passes the $\phi_d$, $\theta_d$, and $U_{\Sigma}$ from the PC to the attitude controller. 
TWCC operates at 10 Hz, the same frequency as GNSS measurements.

\begin{figure}[h]
    \centering
    \includegraphics[scale=0.8]{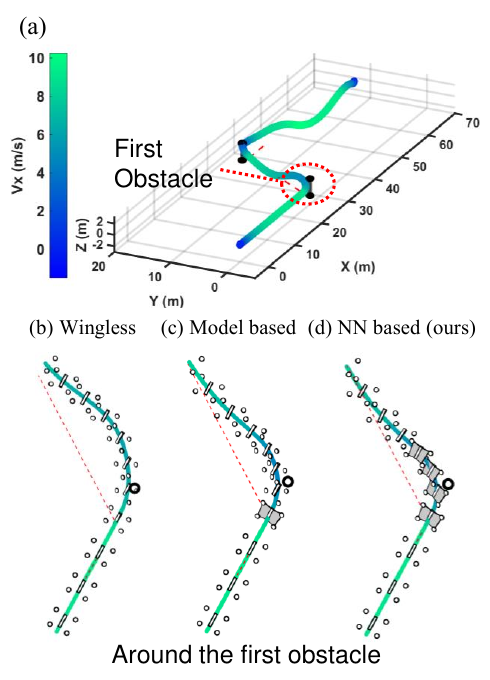}
    \caption{3D and projected 2D flight trajectories of a drone (a) Trajectory for the proposed neural network(NN) based method (b)(c)(d) Projected 2D trajectories for the wingless pure thrust control method, the model based one, and the proposed NN based one. }
    \label{APSC_simulate_result}
\end{figure}

Substituting $\theta_d,~\phi_d,~U_{\Sigma}$ from WIIC into the motion dynamics (\ref{equation of motion}) of the drone yields the same result as substituting $\theta_d,~\phi_d,~U_{\Sigma}$ generated by the PC without spreading the foldable wings. This implies that stability in the $x$- and $y$-directions is preserved. Furthermore, it indicates that by spreading the foldable wings and switching the controller to WIIC, the system can handle cases where PC demands $\theta_d,~\phi_d,~U_{\Sigma}$ beyond the allowable range, which would otherwise compromise stability. Here, WIIC does not account for $z$-axis tracking errors, as lateral motion is more power-efficient than vertical motion when avoiding sudden vertical obstacles.
In Fig. \ref{APSC_simulate_result}, the difference in trajectory tracking performance is compared based on the presence or absence of foldable wings and the method used to calculate aerodynamic forces of wings in TWCC. The aerodynamics force can be calculated using two methods; One is model-based on flat wing theory and the other is data-driven using neural networks. Results show that the latter achieves better tracking performance, especially in cornering, than the former. The process of learning the aerodynamic force generated by the wings will be detailed in the following section.


\section{Learned Aerodynamics of Silicone Wings  \label{section4}}
This work utilizes silicone-based wing membranes to generate additional thrust, improving the tracking performance of a flying squirrel drone. Since silicone allows the wings to deform freely based on speed and attitude, explicitly modeling the resulting air resistance is intractable. To address this, we adopt a data-driven learning approach to learn wing aerodynamics and enhance drone control.

Wind tunnel experiments typically involve a drone with a fixed position but free attitude \cite{cigada2001development,gerges2003wind}. While easy to conduct, this setup bakes it difficult to accurately capture the interaction between air resistance and the drone's state (e.g., pose, velocity). Measuring air resistance in dynamic conditions is even more challenging due to turbulence. Thus, flying drones to collect data is more practical for real-world applications.

We collected the training data in a manner similar to the one described in \cite{o2022neural}. To capture a variety of drone dynamics, its flight is randomly performed with its wings spread. At this time, the drone's dynamics (\ref{equation of motion}) can be written by receiving air resistance \( \bold{f}_a \) as input, as follows:
\begin{equation}\label{eqn_for_label}
\mathbf{f}_a = m\ddot{\mathbf{r}}-mg\mathbf{e}_3 - R_{wb}\mathbf{f} 
\end{equation}
According to the flat wing theory in (\ref{flat theory in 3D_1}), the partial state variables \([ \dot{x}, \dot{y}, \dot{z}, \phi, \theta, \psi ]\) and the corresponding aerodynamic forces are obtained and paired, which will be used as inputs and labels for supervised learning. While the drone performs random maneuvers, the resulting data is collected through GNSS, IMU, and a barometer, including \([ \ddot{x}, \ddot{y}, \ddot{z}, \dot{x}, \dot{y}, \dot{z}, \phi, \theta, \psi, U_\Sigma ]\). The variables \([ \dot{x}, \dot{y}, \dot{z}, \phi, \theta, \psi ]\) are used as input data for supervised learning, and  the aerodynamic forces \( \bold{f}_a \), which serve as the label for learning, can be determined by substituting \([ \ddot{x}, \ddot{y}, \ddot{z}, \phi, \theta, \psi, U_\Sigma ]\) into equation (\ref{eqn_for_label}). For efficient learning of the aerodynamic forces generated by the wings, the following additional considerations are made:
\begin{itemize}
\item Since the wing membrane deforms based on the applied force, its aerodynamic characteristics vary depending on the magnitude and duration of the acceleration force. Therefore, Recurrent Neural Networks (RNNs) are more suitable for learning the aerodynamic forces of wings than general neural networks.

\item  In general flight, the available training data is often insufficient. Most of the time, the drone's desired direction of travel opposes the resulting air resistance, making the data less useful. Useful data is generated only when they align, typically during brief moments when the drone stops or changes direction. This scarcity of useful data can lead to overfitting when learning aerodynamics.


\end{itemize}
To reflect the above considerations, a physics-assisted RNN (paRNN) is proposed, which effectively incorporates a priori physical knowledge and relies on supervised learning only for the remaining hard-to-model dynamics.


Theoretically, the aerodynamic force in (\ref{flat theory in 3D_1}) acts orthogonally to the corresponding plate. However, since the silicone-based wing deforms under airflow, this formula may not hold. Experiments by \cite{maqsood2012flexible} show that aerodynamic force varies with wing material, mounting method, and angle of attack (AOA). Similarly, \cite{breukels2013aeroelastic} demonstrated that wing curvature significantly affects aerodynamic force, causing deviations from theoretical estimates. These findings suggest that aerodynamic force orientation continuously shifts, diverging from flat-wing theory predictions.


\begin{figure}[h]
    \centering
    \includegraphics[width=0.8\linewidth]{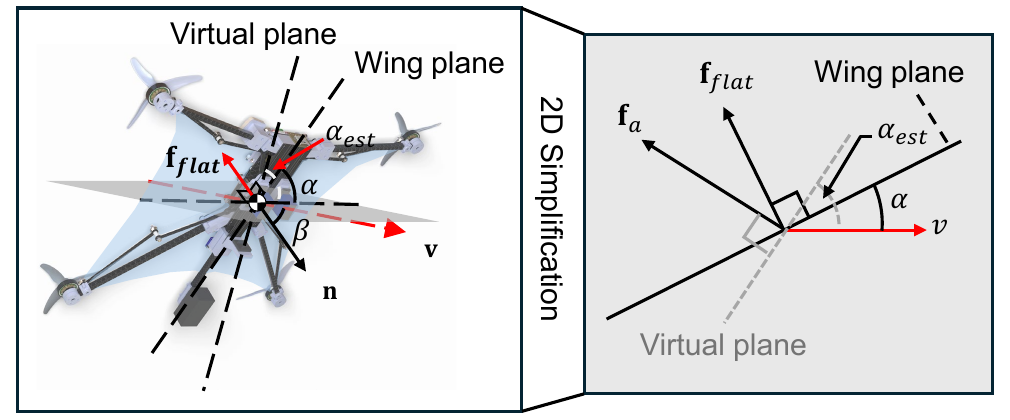}
    \caption{A virtual plane and its corresponding estimated AOA,  $\alpha_{est}.$}
    \label{virtual plane}
\end{figure}
Fig. \ref{virtual plane} compares the theoretically estimated aerodynamic force $\mathbf{f}_{flat}$ with the true aerodynamic force $\mathbf{f}_{a}$ for a drone flying at a constant speed with its wings spread. Here, a virtual plane is orthogonal to $\mathbf{f}_{a}$ has a corresponding AOA denoted as $\alpha_{est}$.
In this work, the aerodynamic force given by (\ref{flat theory in 3D_1}) is estimated by replacing $\alpha$ with $\alpha_{est}$ which is output by paRNN. The effective aerodynamic force is scaled by $\gamma$, which is output by paRNN.

\begin{figure}[h]
    \centering
    \includegraphics[width=0.8\linewidth]{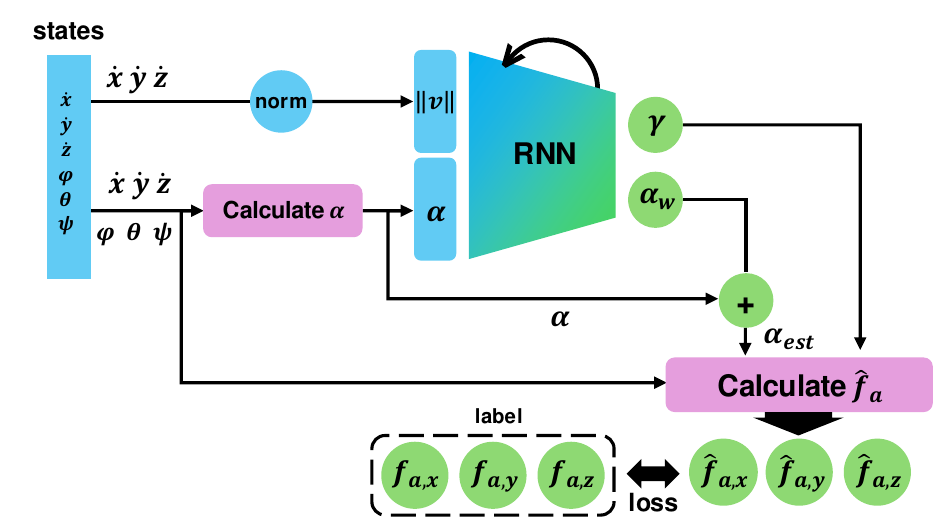}
    \caption{RNN-based learning of air resistance from state variables of a drone.}
    \label{pcRNNs_pipeline}
\end{figure}
Fig. \ref{pcRNNs_pipeline} schematically illustrates the aerodynamic force estimation process, specifically highlighting an RNN that generates $\gamma$ and $\alpha_{w}$. From the RNN outputs, the magnitude and the direction of the estimated air resistance $\hat{\mathbf{f}}_a$ can be determined. Its magnitude. $f_{size}$, is given by 
\begin{align}
f_{size} = \sin\alpha_{est}\rho \left\|\mathbf{v}\right\|^2 A \gamma \label{f_size}
 \end{align}
where $\alpha_{est}$ is given by  
\begin{align}
\alpha_{est} = {\rm{clip}}(\alpha + \alpha_w, \min=-\pi/2, \max=\pi/2) 
 \end{align}
The direction of $\mathbf{f}_a$, $\mathbf{n_{f}}$, can be written as
\begin{eqnarray}
\mathbf{n_{f}} = \mathbf{M^{-1}b}
\end{eqnarray}
where $\mathbf{M}$ and $\mathbf{b}$ are given by
\begin{gather}\label{neural_aeroforce:normal}
\mathbf{M} = 
\begin{pmatrix}
\mathbf{n_v^T}\\
\mathbf{n^T}\\
\mathbf{(n_v \times n)^T}
\end{pmatrix}
, \; \mathbf{b} = \begin{pmatrix} 
\cos(\frac{\pi}{2}-\alpha_{est})\\
\cos\alpha_w\\
0
\end{pmatrix}
\end{gather}
where $\mathbf{n}$ defined by (\ref{flat theory in 3D_3}) and $\mathbf{n_v}$ is parallel to the translational speed $\mathbf{v}$. $\mathbf{n}$ and $\mathbf{n_v}$ are unit vectors. Combining the magnitude and direction, we obtain following equation.
\begin{equation} \label{neural_aeroforce}
\mathbf{\hat{f}}_a = f_{size} \cdot \mathbf{n_f}
\end{equation} 
Note that $f_{size}$ and $\mathbf{n_f}$ in (\ref{neural_aeroforce}) can be computed from the two outputs of paRNN, namely $\alpha_{w}$ and $\gamma$. Compared to directly estimating $\mathbf{f}_a$, this method is more robust to issue such as noise and bias with a small amount of data. Furthermore, the designed RNN is lightweight enough to be easily deployed on a microprocessor and run in real-time. It uses sequence length of 10 and has two hidden layers with 8 nodes each. Training is conducted over 800 episodes. RNN inference runs at 10Hz, the same frequency as the drone's position control.

\section{Experiments \label{section5}}

\subsection{Effectiveness of paRNN}
\begin{figure}[ht]
    \centering
    \includegraphics[width=\linewidth]{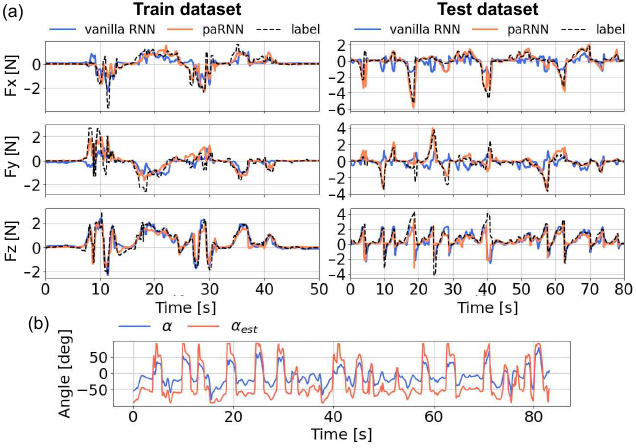}
    \caption{ (a) shows how well the vanilla RNN and our proposed paRNN estimate aerodynamics. (b) illustrates how the $\alpha_{est}$ generated internally by paRNN changes compared to the actual $\alpha$ during the aerodynamic estimation. }
    \label{silicone_wing}
\end{figure}
To validate paRNN, a vanilla RNN for direct estimation of $\mathbf{f_a}$ is employed for comparison. For fairness, both of two take  $\left\| \mathbf{v} \right\|$ and AOA($\alpha$) as inputs. The only difference is that the vanilla RNN directly outputs the 3D aerodynamic force without estimating any parameter values such as $\alpha_w$ and $\gamma$. In Fig. \ref{silicone_wing}(a), the performance of paRNN is better than that of the vanilla RNN, specially for test datasets. It means that paRNN avoids overfitting and improves the sample efficiency. The quantitative comparison results are shown in Table \ref{table:eval model}.
\begin{table}[h]
\caption{Aerodynamic Estimation Errors (RMSE)}
\label{table:eval model}
\begin{center}
\begin{tabular}{|c|c|c|c|}
\hline
\textbf{dataset} & \makecell{speed range [m/s]} & \makecell{vanilla RNN} & \makecell{paRNN (ours)} \\
\hline
\makecell{trainset} & \makecell{[0.3080,7.9472]} & \makecell{0.4821[N]}  & \makecell{\textbf{0.4209[N]}}\\
\hline
\makecell{testset 1} & \makecell{[0.6697,10.677]} & \makecell{0.9375[N]}  & \makecell{\textbf{0.6965[N]}}\\
\hline
\makecell{testset 2} & \makecell{[0.1936,7.4728]} & \makecell{0.6977[N]}  & \makecell{\textbf{0.4094[N]}}\\
\hline
\makecell{testset 3} & \makecell{[0.2503,7.9532]} & \makecell{0.9672[N]}  & \makecell{\textbf{0.6019[N]}}\\
\hline
\makecell{testset 4} & \makecell{[0.3489,5.7032]} & \makecell{0.3133[N]}  & \makecell{\textbf{0.2518[N]}}\\
\hline
\end{tabular}
\end{center}
\end{table}

\noindent Fig. \ref{silicone_wing}(b) shows short sequence of evaluation of  $\alpha_{est}$ from paRNN. It is noted that paRNN consistently produces $\alpha_{est}$ greater in magnitude
than $\alpha$, to correct the increased aerodynmic force from morphed silicone wing.

\subsection{Trajectory Generation}
\begin{figure}[ht]
    \centering
    \includegraphics[width=0.8\linewidth]{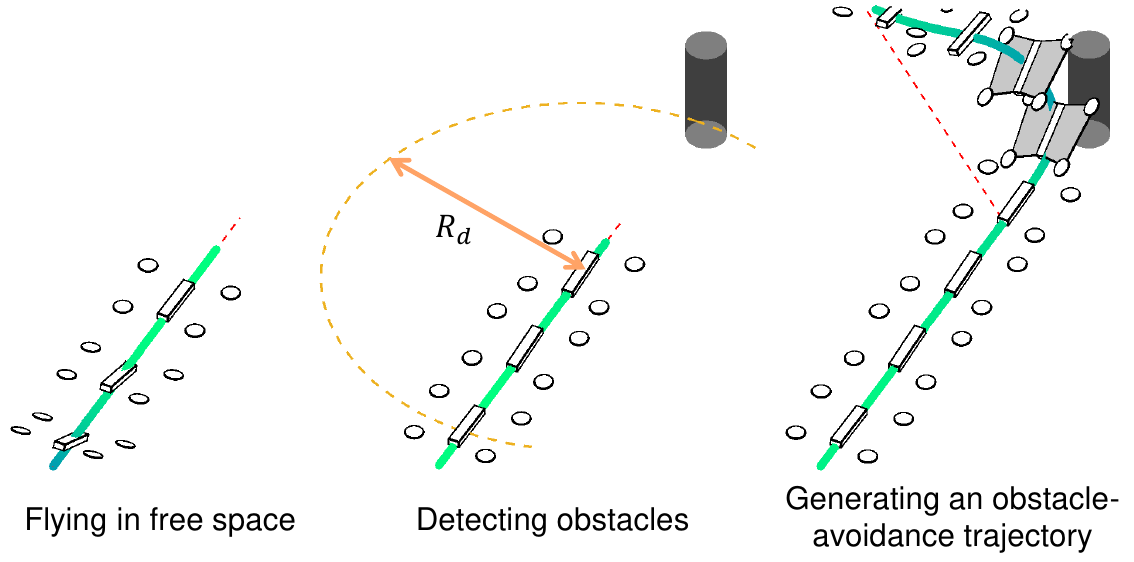}
    \caption{Trajectory generation process.}
    \label{trajectory generation}
\end{figure}
To assess how effectively TWCC utilizes aerodynamic forces, we examine how to avoid tall obstacles at high speeds. A drone detects obstacles within a sensing range of \( R_d \) and, upon detection, plans a new avoidance trajectory. Specifically, obstacles within \( R_d \) and a \( \pm 15^\circ \) field of view in the direction of movement of the drone are detected. The maximum roll and pitch angles are set to \( \phi_{max} = 30^\circ,~\theta_{max} = 30^\circ \) for both simulation and experiment.
A reference trajectory is generated using a 3-rd order polynomial until the target speed is reached, and then a 1st order polynomial is used to maintain a constant speed. Obstacle locations are defined relative to the starting point of a newly generated trajectory. The drone measures its distance to an obstacle via GNSS, which also serves as a detection criterion.
This experiment compares tracking performance in maneuvering scenarios requiring sudden acceleration rather than in optimal trajectories for avoidance. As computations must be performed in real time on an onboard system, we employ a simple trajectory generation method instead of an advanced complex one. Fig. \ref{trajectory generation} illustrates the trajectory generation process.
\subsection{Reference Trajectory Following}
The methodology proposed in this study was validated through simulations and real-world experiments. Both in simulations and real-world experiments, we compare a wingless drone, a flat wing model-based TWCC, and the proposed paRNN-based TWCC. 

\begin{figure}[b]
    \centering
    \includegraphics[width=\linewidth]{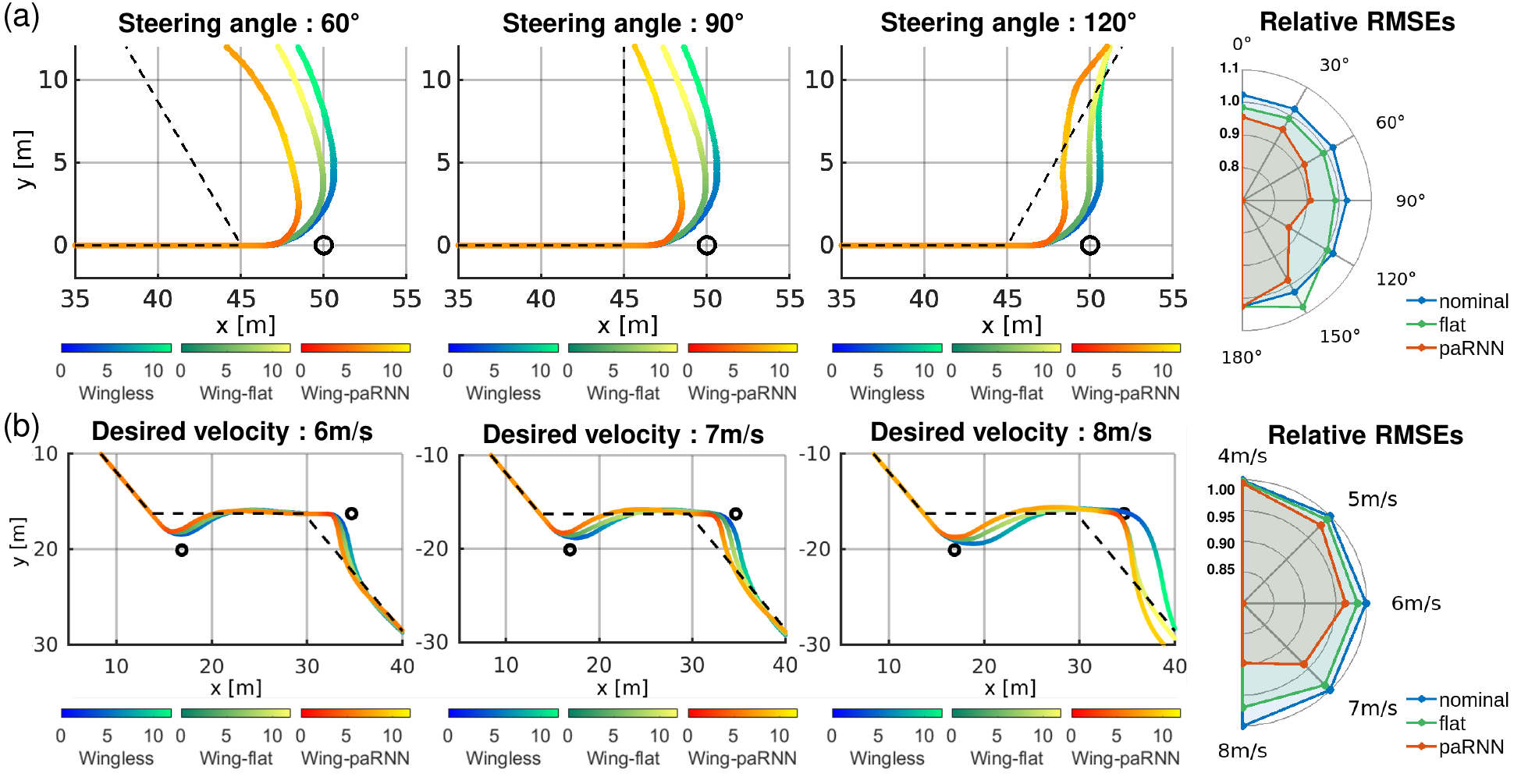}
    \caption{  (a) and (b) show the tracking trajectories for different steering angles and desired velocities, respectively.
The unit of colorbar is m/s. The rightmost circular plots represent the ratio of each RMSE relative to that of a conventional controller for a wingless drone. 'Wingless', 'Wing model-based', and `paRNN' denotes a TWCC controller for a wingless drone, a flat wing model based controller, and the proposed paRNN based controller, respectively.}
    \label{RMSE_by_angle}
\end{figure}

\vspace{0.6em}\noindent \textbf{Simulation Experiments.}
Fig. \ref{RMSE_by_angle} shows simulation results for different steering angles and desired velocities. Aerodynamic simulation is performed using paRNN, with $R_d=5$ in the simulation.
As shown in the rightmost plots of Fig. \ref{RMSE_by_angle}(a), the proposed paRNN-based control scheme achieves the best performance in all steering directions and desired velocities. As the desired velocity increases, the effectiveness of silicone wings improves, enhancing the performance of the proposed paRNN-based control scheme.

\begin{figure}[h]
    \centering
    \includegraphics[width=0.7\linewidth]{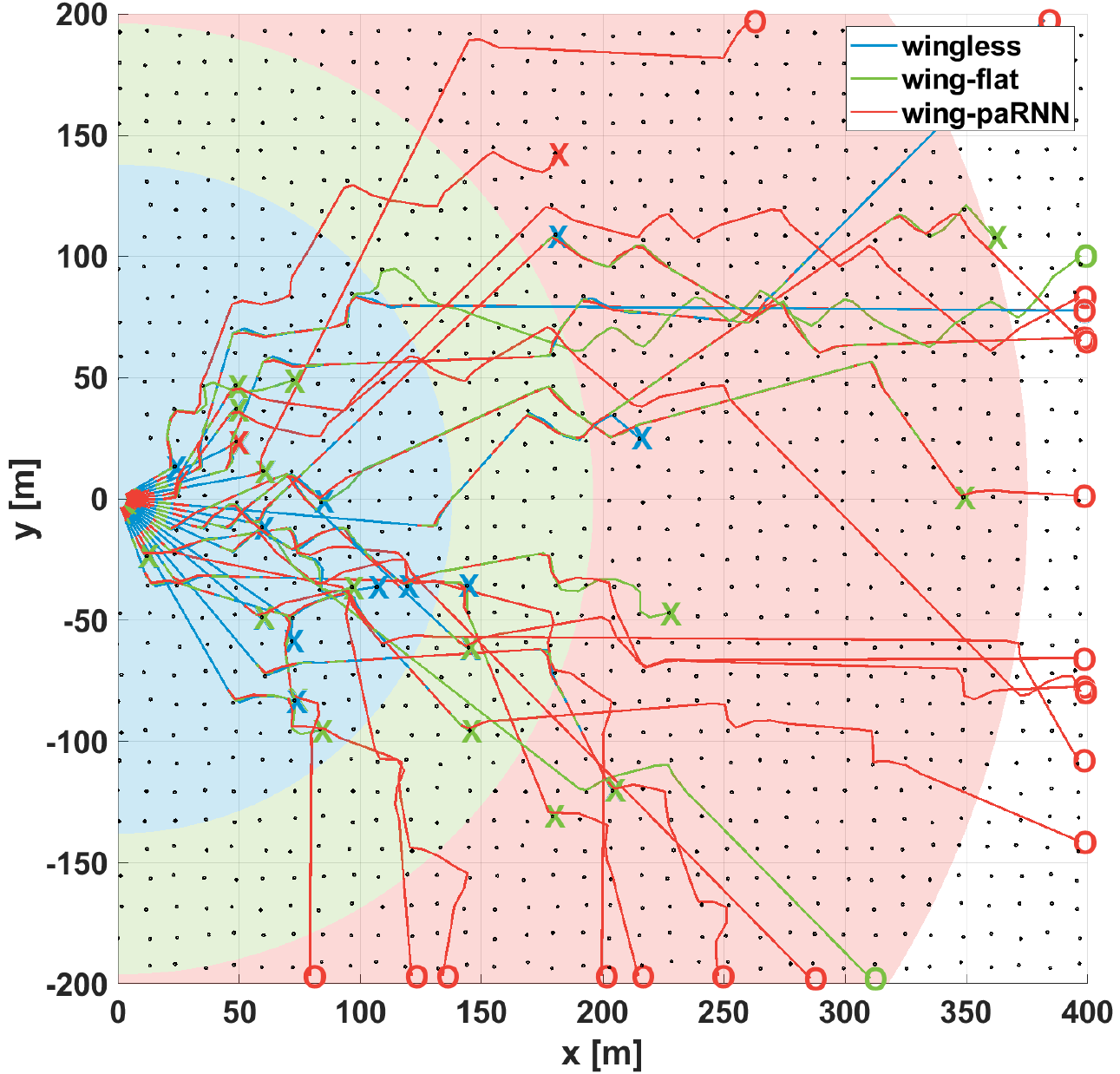}
    \caption{Simulation results in a forest-like environment. The locations where the drone collided with an obstacle are marked with ``X''. Drones that successfully navigated through the forest without collisions are marked with ``O''. The forest size is 400 × 400 m, and each obstacle has a diameter of 1m. Small dots spread across the plot represent obstacles. The average travel distance of the drones, measured from the origin, is represented as a shaded semicircles on the graph. (Collision-free drones flew over 400 m.)}
    \label{largescale_sim}
\end{figure}

To obtain more general experimental results, we conducted simulations in a forest-like environment with obstacles placed at regular intervals. The trajectory generation method remained the same as in Section V.C, while the steering angle was randomly selected between the range of [110,130] degrees whenever an obstacle was encountered. Obstacle spacing was randomly set within the range of [10.5,13.5] m. The desired velocity was 7 m/s. The results are shown in Fig. \ref{largescale_sim}.
Each controller was tested in 21 trials. The paRNN-based TWCC (red) successfully navigated through the forest in 90.5\% of trials, colliding only twice. In contrast, the wingless drone (blue) and the flat wing model-based TWCC (green) had significantly lower success rates, completing only 9.52\% and 19.0\% of the trials, respectively. 
A comparison of average travel distances confirms that the paRNN-based TWCC outperforms the other two controllers.
\begin{figure}[b]
    \centering
    \includegraphics[width=0.8\linewidth]{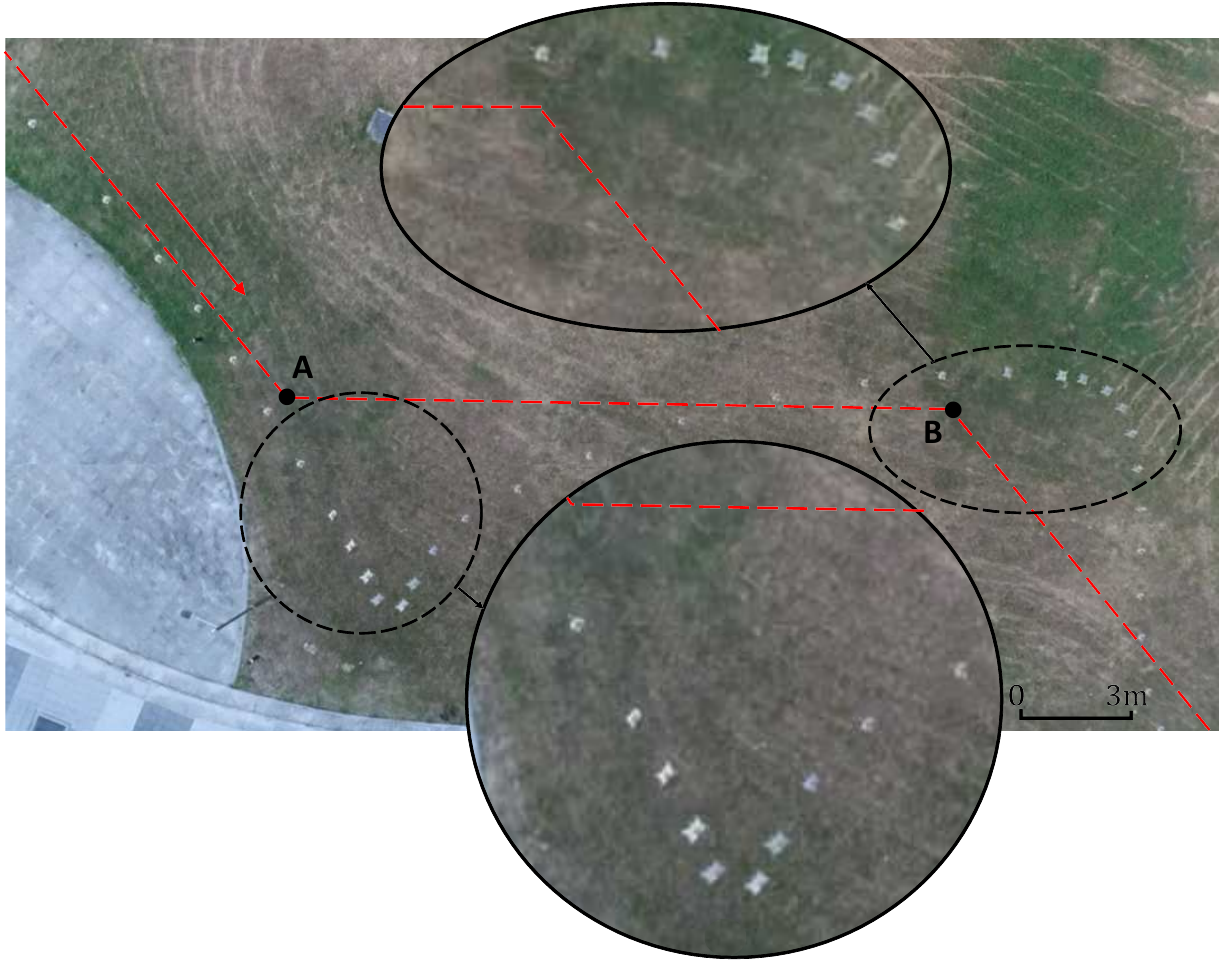}
    \caption{Real world experiment.}
    \label{exp_image_latex}
\end{figure}
\begin{figure}[t]
    \centering
    \includegraphics[width=\linewidth]{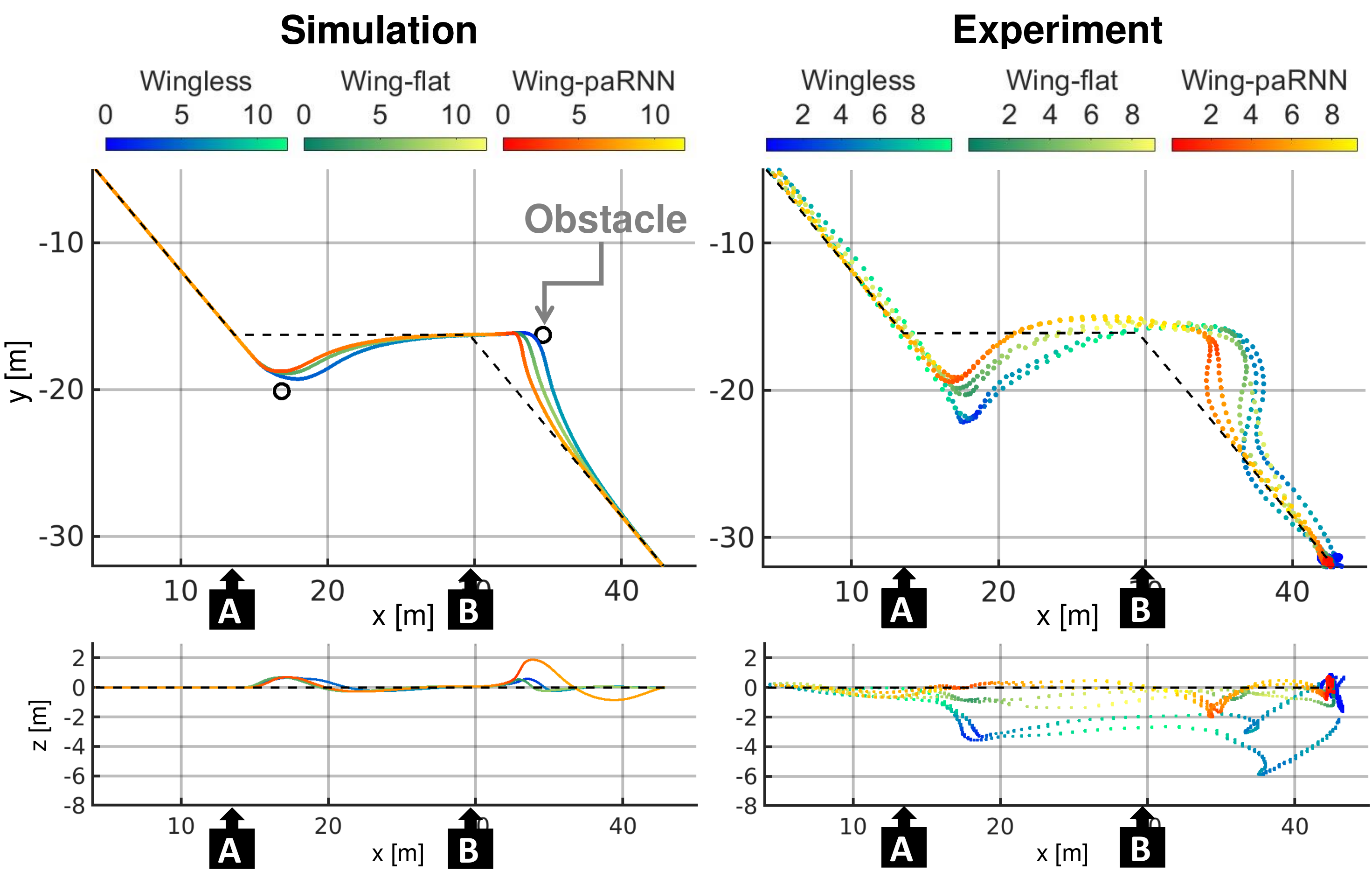}
    \caption{Comparison of tracking trajectories between simulations and real-world experiments. The unit of colorbar is m/s. The experimental graph plots position data obtained through GNSS for the x and y axes, while the z-axis data is derived from barometer and IMU using Kalman filtering.}
    \label{skyview_2}
\end{figure}

\vspace{0.6em}\noindent\textbf{Real World Experiments.} The desired velocity was set to 7.3 m/s during real-world experiments, representing the fastest stable flight achievable for the experimental drone. The experiment was conducted under the same conditions as the previous simulation, except that $R_d$ was set to 4 m.
As shown in Fig. \ref{skyview_2}, the proposed paRNN-based wing control scheme significantly improves tracking performance.
Notably, unlike in the simulation, a conventional wingless drone tends to sink in the $z$-direction while detecting and avoiding obstacles. This behavior is hypothesized to result from the drone's battery, which has a maximum C-rate of 75C, potentially preventing sufficient thrust generation during highly agile maneuvers.
In contrast, the flying squirrel drone, equipped with the proposed control scheme, generates additional force in the $z$-direction by spreading its wings, preventing it from sinking. 
In real-world experiments, the difference in trajectory tracking performance between the wingless drone and the drone with foldable wings was more pronounced than in simulation results. This discrepancy is presumed to be due to the lower performance of the wingless drone in real-world experiments compared to simulations, likely caused by sim-to-real gaps such as the battery's C-rate limitations, as mentioned earlier.

\begin{table}[ht]
\caption{trajectory tracking RMSE in real world}
\label{table:real demo}
\begin{center}
\begin{tabular}{|c|c|c|c|}
\hline
\textbf{Trial} & \makecell{Wingless} & \makecell{ Wing model-based} & \makecell{paRNN} \\
\hline
\makecell{1} & \makecell{4.7012[m]}  & \makecell{4.6432[m]} & \makecell{\textbf{3.9292[m]}}\\
\hline
\makecell{2} & \makecell{4.4438[m]}  & \makecell{4.3026[m]} & \makecell{\textbf{4.1752[m]}}\\
\hline
\makecell{3} & \makecell{4.8829[m]}  & \makecell{4.5007[m]} & \makecell{\textbf{4.2986[m]}}\\
\hline
\multicolumn{4}{l}{\parbox[t]{0.4\textwidth}{$^{\mathrm{a}}$‘Wingless’, ‘Wing model-based’, and ‘paRNN’ represent the same meanings as in Fig. 9.}}
\end{tabular}
\end{center}
\end{table}

\section{Conclusions and Futureworks \label{section6}}
This paper presented a strategic and cooperative approach to controlling the silicone wings of a flying squirrel drone. The complex aerodynamics of the silicone wings were effectively modeled using a recurrent neural network (RNN) and integrated into a controller to enhance tracking performance. The resulting improvements were significant, enabling the drone to avoid real-world crashes that are typically unavoidable for wingless drones. This work highlights the effective application of biomimicry, leveraging the dynamics of living organisms to precisely regulate the agile movements of drones.

The proposed data-driven approach could be further improved by incorporating efficient trajectory design, offering a promising direction for future research.

\addtolength{\textheight}{-5cm}   





\bibliographystyle{IEEEtran}
\bibliography{ref.bib}

\end{document}